\documentclass{fairmeta}
\usepackage{amsmath}
\usepackage{amssymb}
\usepackage{amsfonts}
\usepackage{nicefrac}
\usepackage{url}
\usepackage{enumitem}
\usepackage{float}

\newcommand{\ours}{CompassAD}
\newcommand{\task}{Intent-Driven Confusable Affordance Grounding}
\newcommand{\ici}{ICI}
\newcommand{\bcr}{BCR}
\newcommand{\model}{CompassNet}

\newcommand{\loss}{\mathcal{L}}

\providecommand{\vs}{vs.}

\newcommand{\authorlink}[2]{\href{#1}{\color{black}#2}}

\renewcommand\authorformat[2][]{{\sffamily\bfseries #2\ifx\\#1\\\else$^{#1}$\fi}}
\renewcommand\affiliationformat[2][]{{\normalsize\ifx\\#1\\\else$^{#1}$\fi#2}}

\title{CompassAD: Intent-Driven 3D Affordance Grounding in Functionally Competing Objects}

\author{\authorlink{https://lorenzo-0-0.github.io/}{Jingliang Li}}
\author{\authorlink{https://jiajindou.github.io/}{Jindou Jia}}
\author{\authorlink{https://morpheus-an.github.io/}{Tuo An}}
\author{\authorlink{https://chuhaozhou99.github.io/Chuhao-Zhou/}{Chuhao Zhou}}
\author{\authorlink{https://scholar.google.com/citations?user=wEV8pmkAAAAJ}{Xiangyu Chen}}
\author{\authorlink{https://shanshilin.github.io/}{Shilin Shan}}
\author{\authorlink{https://ma-boyu.github.io/}{Boyu Ma}}
\author{Bofan Lyu}
\author[\dagger]{\authorlink{https://reagan1311.github.io/}{Gen Li}}
\author[\dagger]{\authorlink{https://marsyang.site/}{Jianfei Yang}}

\affiliation{MARS Lab, Nanyang Technological University, Singapore}

\contribution[\dagger]{Corresponding authors}

\correspondence{Jianfei Yang (\email{jianfei.yang@ntu.edu.sg}); Gen Li (\email{gen.li@ntu.edu.sg}); Jingliang Li (\email{jinglian001@e.ntu.edu.sg})}

\metadata[Project Page]{\href{https://lorenzo-0-0.github.io/CompassAD-Intent-Driven-3D-Affordance-Grounding-in-Functionally-Competing-Objects/}{lorenzo-0-0.github.io/CompassAD}}

\abstract{
When told to ``cut the cake,'' a robot must choose the knife over nearby scissors, despite both objects affording the same cutting function.
In real-world scenes, multiple objects may share identical affordances, yet only one is appropriate under the given task context.
We call such cases confusing pairs.
However, existing 3D affordance methods largely sidestep this challenge by evaluating isolated single objects, often with explicit category names provided in the query.
We formalize \task{}, a new 3D affordance setting that requires predicting a per-point affordance mask on the correct object within a multi-object point cloud, conditioned on implicit natural language intent.
To study this problem, we construct \ours{}, the first benchmark centered on implicit intent in confusing multi-object compositions. It comprises 30 confusing object pairs spanning 16 affordance types, 6,422 compositions, and 88K+ query-answer pairs.
Furthermore, we propose \model{}, a framework that incorporates two dedicated modules tailored to this task. Instance-bounded Cross Injection (\ici{}) constrains language-geometry alignment within object boundaries to prevent cross-object semantic leakage. Bi-level Contrastive Refinement (\bcr{}) enforces discrimination at both geometric-group and point levels, sharpening distinctions between target and confusable surfaces.
Extensive experiments demonstrate state-of-the-art results on both seen and unseen queries, and deployment on a robotic manipulator confirms effective transfer to real-world grasping in confusing multi-object compositions.
}

\begin{document}

\maketitle

\section{Introduction}
\label{sec:intro}

Real-world environments are inherently cluttered: kitchens, workshops, and offices all present multiple objects in close proximity, often in disorganized arrangements.
For an embodied agent operating in such spaces, understanding what each object affords~\cite{gibson1979ecological}, i.e., the functional interactions it supports, is a prerequisite for safe and effective manipulation.
3D affordance research has made encouraging progress toward this goal, yet the vast majority of existing work~\cite{deng2021iad,yang2023iagnet,tai2024great,li2024laso,xu2024glance} studies affordance in a controlled single-object setting, localizing functional regions on one isolated item at a time.
Real-world agents, however, must perceive and reason over entire multi-object compositions rather than sanitized individual instances, making the extension from single-object to multi-object affordance grounding far from trivial.
% Nonetheless, extending affordance grounding from one object to many is far from a trivial increase in quantity.
A natural direction is to move toward full 3D scene-level affordance understanding~\cite{delitzas2024scenefun3d, liu2024grounding, li2025seqaffordsplat}.
% : recent scene-level benchmarks such as SceneFun3D~\cite{delitzas2024scenefun3d, liu2024grounding, li2025seqaffordsplat} annotate fine-grained functional elements (e.g., handles, knobs, switches).
However, this setting introduces substantial computational overhead and dense background geometry, which can distract the model and degrade grounding performance.

% ============================================================
% TEASER FIGURE (floats to top of next page via [!t])
% ============================================================
\begin{figure}[!t]
\centering
\includegraphics[width=\textwidth]{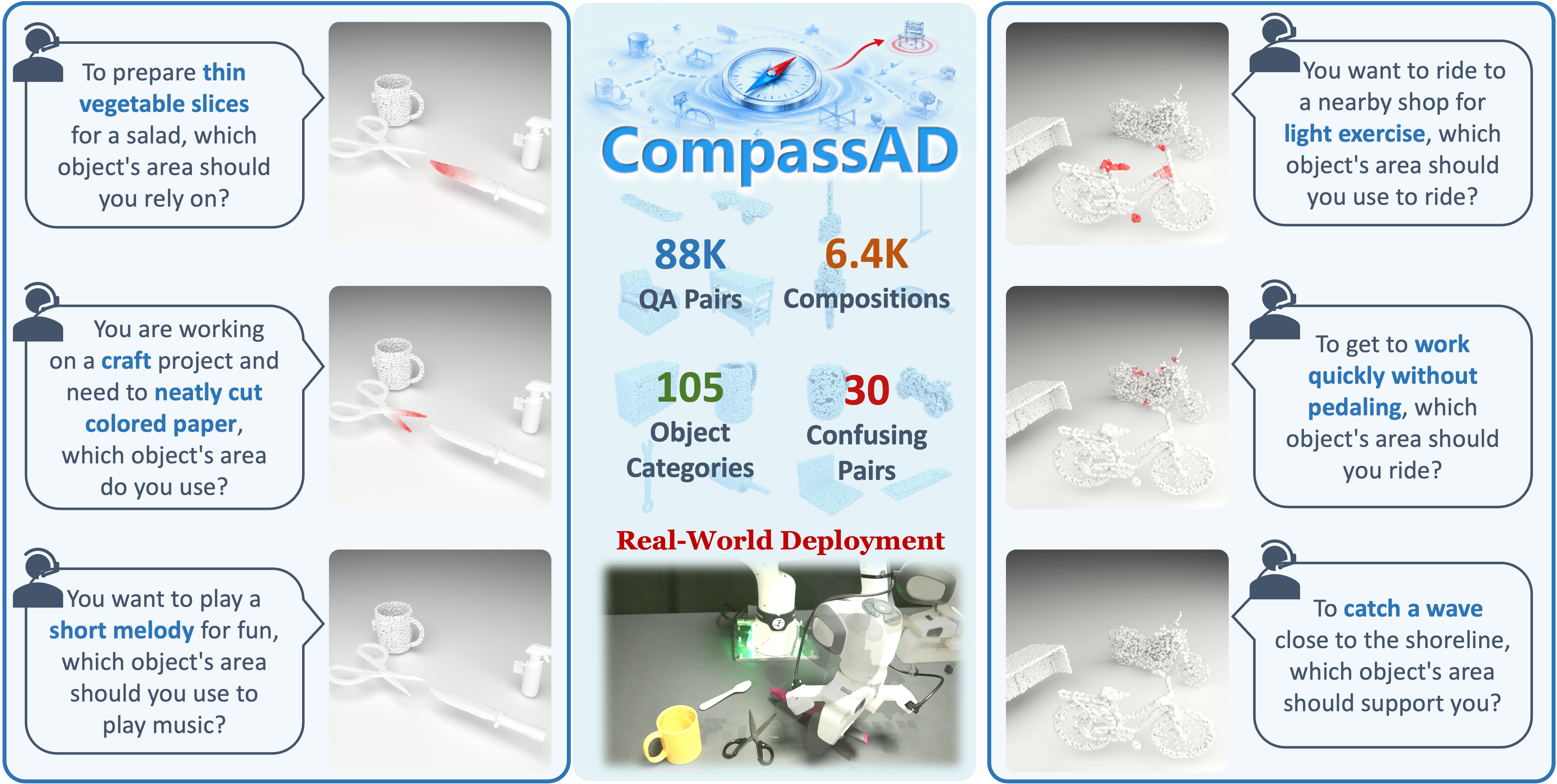}
\caption{\textbf{The proposed task of \task{}.} Given a multi-object point cloud and a natural language query describing an intended action, the goal is to predict a per-point affordance mask. The same composition can yield different targets depending on the query intent.
We introduce \ours{}, a benchmark for this challenging setting.}
\label{fig:teaser}
\end{figure}

This leaves a critical middle regime underexplored: cluttered local workspaces where objects with overlapping functional roles naturally cluster together. We term such co-occurring objects \emph{confusing pairs}: distinct object categories that support the same target affordance and play similar functional roles, sometimes accompanied by similar visual or geometric characteristics. For instance, a knife and a pair of scissors both afford cutting despite their markedly different geometries, while a skateboard and a surfboard share the riding affordance with highly identical structural designs. In either case, the shared affordance renders the target ambiguous from appearance alone, and correct selection must therefore be conditioned on task intent.

% while each instruction is curated to refer to a unique functional element.
%This leaves a critical middle regime underexplored: cluttered local workspaces where objects sharing similar functional properties naturally cluster together.
%For example, a kitchen counter often hosts a cup beside a bowl, both affording containment; knives and scissors frequently appear together in toolboxes, both affording cutting.
%Such co-occurring objects, which we term confusing pairs, share identical affordance types.
%Correct selection must therefore be conditioned on task intent.
%However, human instructions are typically object-agnostic and implicitly expressed. Rather than explicitly naming the object (e.g., ``find the cup that can contain coffee''), people communicate intent in a goal-driven manner (e.g., ``I would like to have some coffee'').
%This implicit, intent-driven nature further complicates target selection.

Together, functional overlap and implicit intent make multi-object affordance reasoning inherently query-dependent: the same composition can yield different affordance regions depending on the instruction.
Consider a scissors-knife pair placed on a table (\cref{fig:teaser}).
For the query ``I need to prepare vegetable slices,'' the knife blade should be activated; for ``I want to cut paper,'' the scissors' blade should be highlighted.
In contrast, if the intent is to play music, no activation should occur, as neither object affords that function.
This query-dependent nature, combined with diverse distractors, precludes spatial memorization and poses a challenge that single-object methods struggle to solve.

These observations raise a central question: \emph{Can we achieve 3D affordance grounding in confusing multi-object compositions under implicit intents?}
We formalize this task as \textbf{\task{}}: given a multi-object point cloud containing confusing pairs and a natural language query describing functional intent, the goal is to predict a per-point affordance mask on the correct object, disambiguating confusable alternatives and abstaining when appropriate.
Existing 3D affordance benchmarks~\cite{deng2021iad,li2024laso} primarily focus on single objects and often include target object names in the instruction.
To address this gap, we construct \textbf{\ours{}}, the first 3D affordance benchmark designed around implicit intent in confusable multi-object compositions.
\ours{} comprises 30 confusing object pairs spanning 16 affordance types, 6,422 compositions, and over 88K query-answer pairs, all with query-dependent ground truth.
The dataset emphasizes target-free instructions, unseen query generalization, and abstention testing to ensure safe behavior.

When naively adapted to \ours{}, state-of-the-art single-object methods~\cite{li2024laso,xu2024glance} expose two systematic failure modes that existing benchmarks rarely reveal.
First, global cross-modal fusion diffuses query semantics across instances, so activations frequently land on the wrong object when a distractor shares the queried affordance type.
Second, even when the correct instance is reached, the predicted affordance region remains coarse and imprecise, and queries that no object can satisfy still trigger confident activations rather than correct abstention.
%Second, even when the correct instance is reached, predictions leak onto geometrically similar surfaces of competing objects, and queries that no object can satisfy still trigger confident activations rather than correct abstention.
Together, these failures reveal that \task{} cannot be reduced to scaling existing methods to larger inputs: it requires joint resolution of intent-driven instance selection and fine-grained point localization, and an error at either level invalidates the prediction.
%This diagnosis motivates two complementary principles.
%At the coarse scale, cross-modal fusion should be instance-bounded, so that language selects among confusing objects without diffusing across object boundaries.
%At the fine scale, the model requires explicit discriminative supervision against high-confidence false positives.

To this end, we propose \textbf{\model{}}, which couples two complementary modules along a shared instance--region--point hierarchy.
Rather than performing unrestricted cross-modal fusion over the whole point cloud, Instance-bounded Cross Injection (\ici{}) confines language--geometry interaction within each object boundary, eliminating cross-instance semantic leakage at its source while preserving fine-grained localization.
Bi-level Contrastive Refinement (\bcr{}) then layers two complementary training-only signals on top: Region-level contrast sharpens the selection of the object whose functional regions best match the intent, while point-level hard negative supervision suppresses high-confidence false activations on competing objects or similar surfaces. Notably, \bcr{} is applied only during training and introduces no additional inference cost.
%Instance-bounded Cross Injection (\ici{}) keeps cross-modal interaction within each object: points are grouped into functional regions that attend to the query together with a learnable background token, and the resulting query-aware features are gated back to points. This bounded fusion mitigates the cross-instance leakage that drives wrong-object selection.
%Bi-level Contrastive Refinement (\bcr{}) introduces fine-scale contrastive supervision through two training-only losses. TG-Softmax ranks functional regions by affordance coverage to sharpen object selection, and TP-HardNeg mines high-scoring negatives along confusable surfaces to suppress fine-grained false activations. Both losses share \ici{}'s backbone, so the contrastive gradients refine its forward representations during training; \bcr{} is dropped at inference and adds no overhead.
\model{} outperforms all adapted baselines on \ours{}, improving over the previous best by 28.3\% in aIoU and 24.3\% in SIM. We further validate practical applicability by deploying our method on a robotic manipulator for grasping in confusing multi-object compositions, demonstrating that interpreting implicit intent transfers effectively to real-world manipulation.

Overall, our contributions can be summarized as follows:

\begin{enumerate}[nosep,leftmargin=*]
\item We formalize a novel task of \task{}, where objects with shared affordance types create confusing pairs and correct predictions are query-dependent.

\item We construct \ours{}, the first benchmark tailored to our task, with 30 confusing object pairs across 16 affordance types, 6{,}422 compositions, and 88K+ QA pairs. It includes target-free queries, language generalization splits, and abstention testing for safe decision-making.

\item We propose \model{}, which couples Instance-bounded Cross Injection (\ici{}) for forward cross-modal injection with Bi-level Contrastive Refinement (\bcr{}) for backward contrastive supervision, along a shared instance--region--point hierarchy. This design achieves state-of-the-art results on \ours{} and successful deployment on a real robotic manipulator.
%\item We propose a multi-granularity framework \model{} that integrates two task-specific modules for boundary-aware cross-modal grounding and fine-grained discrimination. Our method achieves state-of-the-art performance on \ours{} and demonstrates successful real-world deployment on a robotic manipulator.
\end{enumerate}

% ============================================================
% 2. RELATED WORK
% ============================================================
\section{Related work}
\label{sec:related}

\noindent\textbf{3D affordance grounding.}
Affordance learning identifies the regions of an object that support physical interactions~\cite{gibson1979ecological,deng2021iad}.
Building on 2D affordance detection~\cite{fang2018demo2vec,nagarajan2019grounded,do2018affordancenet,thermos2020deep,li2023locate,li2024one,li2025learning},
3D AffordanceNet~\cite{deng2021iad} introduced fine-grained affordance labels on point clouds, and later work leveraged visual foundation models~\cite{radford2021learning,qian2024affordancellm,tai2024great,wang2024affogato,chu20253d,nguyen2023open,chen2024worldafford,liu2025pavlm,basakvispla,zhu2025grounding,kim2024beyond,li2025interpretable,wei20253daffordsplat,park2026affostruction,huang2025unlocking} to generalize across categories.
Scene-level work such as SceneFun3D~\cite{delitzas2024scenefun3d,liu2024grounding,li2025seqaffordsplat} extends this to fine-grained functional elements paired with natural-language tasks, but each instruction still targets a unique element.
We instead study the practical case where multiple objects in one composition share the same affordance type, making the correct affordance map inherently query-dependent.

\noindent\textbf{Benchmarks for language-guided 3D affordance grounding.}
Early 3D benchmarks annotate single objects with part-level supervision~\cite{deng2021iad,xu2022partafford}, isolating geometry from compositional context.
Later datasets enrich supervision with interaction evidence from images, egocentric videos, and multi-view cues~\cite{luo2022grounded,luo2022learning,chen2023affordance,bahl2023affordances,liu2024grounding,heidinger20252handedafforder,wang2026videoafford}, while language-guided benchmarks pair free-form queries with per-point masks~\cite{yu2025seqafford,zhu2025grounding,jiang2025affordancesam} under seen/unseen splits.
Yet most~\cite{deng2021iad,xu2022partafford,delitzas2024scenefun3d,song2025learning} still focus on isolated objects or assume a uniquely identifiable target, leaving competing affordances among co-occurring objects unaddressed and conflating object-selection with part-localization errors.
\ours{} introduces a benchmark of controlled confusion in multi-object compositions, where each intent-driven instruction must identify the correct instance and ground its affordance region.

\section{Datasets}
\label{sec:dataset}
\ours{} is designed to evaluate a failure mode that is difficult to isolate in existing 3D affordance benchmarks.
Every composition is built around a curated pair of objects that share the same target affordance, alongside optional distractors that introduce clutter without contributing to affordance ambiguity. Queries are framed in an intent-driven form that never names any object, so disambiguation must be inferred from intent alone. Supervision is tied to the query rather than the composition, and the same composition yields different ground-truth masks under different intents.
%We formalise the task in \cref{sec:task_def}, describe how compositions and queries are constructed in \cref{sec:data_curation}, and analyse key statistics in \cref{sec:dataset_stats}.
%Objects with similar affordances frequently co-occur within real-world environments. This requires a robot to interpret intent-driven queries and select the correct object/region among multiple candidates. Most existing 3D affordance benchmarks~\cite{deng2021iad,li2024laso,xu2024glance} focus on single-object settings. Scene-level datasets (e.g., SceneFun3D~\cite{delitzas2024scenefun3d}) consider full scenes, but do not explicitly model confusing pairs with overlapping functional properties, leaving intent-level ambiguity largely underexplored. To address this gap, we introduce \ours{}, a large-scale benchmark for 3D affordance grounding in confusing multi-object compositions, featuring intent-driven instructions and query-dependent ground truth to systematically evaluate disambiguation under competing functional properties.

\subsection{Task definition}
\label{sec:task_def}
A composition in \ours{} is a multi-object point cloud that approximates the local clutter typical of a real-world workspace. Each composition is assembled from one confusing pair $(o_a, o_b)$ and $D \in \{0,1,2\}$ distractor objects $\{o_d^k\}_{k=1}^{D}$, giving compositions of two to four object instances. The pair establishes the controlled functional confusion that defines the task, while the distractors introduce geometric variety without contributing further confusable candidates.
Each composition is represented as a point cloud $\mathcal{P} = \{p_i\}_{i=1}^{N}$ with $p_i \in \mathbb{R}^3$. Given a natural language query $\mathcal{Q}$ that describes the user's intent, the goal is to learn a model $f_\theta$ that predicts a per-point affordance probability map
\begin{equation}
\hat{\bm{y}} = f_\theta(\mathcal{P}, \mathcal{Q}) \in [0,1]^N,
\end{equation}
where $\hat{y}_i \in [0,1]$ denotes the predicted affordance probability at point $p_i$. The ground truth $\bm{y} \in [0,1]^N$ is query-dependent, meaning that two distinct queries $\mathcal{Q}_a \neq \mathcal{Q}_b$ over the same composition $\mathcal{P}$ may correspond to different masks $\bm{y}_a \neq \bm{y}_b$. To support safe abstention, we additionally include negative queries that no object in $\mathcal{P}$ can satisfy, for which the ground truth is $\bm{y} = \bm{0}$. The task therefore requires the model not only to identify the object referred to by the query but also to localize the corresponding affordance region on its surface.

\subsection{Data curation}
\label{sec:data_curation}

At the core of \ours{} is the notion of a confusing pair, formalized as two object categories that share at least one target affordance and are functionally interchangeable under it, often reinforced by visual or geometric similarity. Following this criterion, 30 pairs spanning 16 affordance types are curated from a diverse object pool that combines synthetic CAD models with real-world scans across multiple established 3D repositories, providing broad geometric and semantic coverage. Each composition is built around one such pair alongside up to two distractor objects drawn from the same pool, with distractors required not to share the pair's target affordance to keep ambiguity strictly pair-level. Compositions are then assembled through a multi-stage pipeline combining per-object normalisation, randomised spatial placement, and permuted slot assignment, which together prevent positional, scale, or contextual shortcuts.

For each composition, we prompt GPT~\cite{achiam2023gpt} under two structured templates yielding affordance-relevant positive queries and affordance-absent negative queries; both templates explicitly forbid any object-category name so that disambiguation must rely on intent. Each candidate first passes an automatic filter rejecting name leakage, out-of-range lengths, and intra-composition duplicates. Surviving queries are then independently reviewed by trained annotators on three explicit criteria, namely intent clarity, target uniqueness for positive queries or unanswerability for negative ones, and natural phrasing, with disagreements resolved by majority vote. GPT is asked to recover the intended target from each query alone, and failure case is returned for regeneration.

\subsection{Statistics and analysis}
\label{sec:dataset_stats}

\begin{figure}[!t]
\centering
\includegraphics[width=0.95\linewidth]{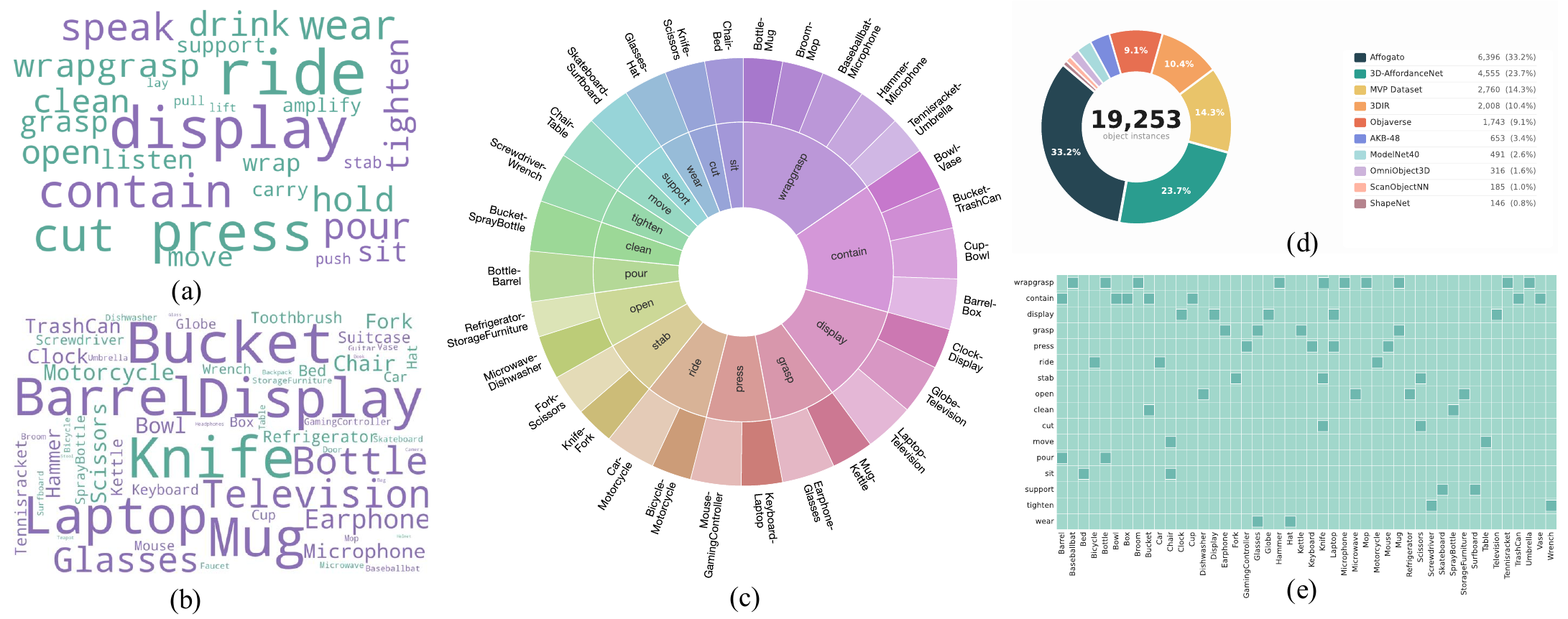}
\caption{\textbf{Overview of the \ours{} benchmark.}
(a) Affordance concept distribution.
(b) Object category distribution.
(c) Hierarchy of confusing pairs grouped by target affordance type.
(d) Source breakdown of the collected 3D object instances.
(e) Confusion matrix between affordance and object categories, highlighting the many-to-many nature of real-world affordances.}
\label{fig:dataset_overview}
\end{figure}

\ours{} comprises 6{,}422 multi-object compositions and 87{,}964 intent-driven language queries that span 105 object categories and 16 target affordance types alongside additional negative-only concepts (\cref{fig:dataset_overview}a--c). The underlying point clouds are sourced from established 3D affordance datasets along with additional synthetic and real-world 3D repositories (\cref{fig:dataset_overview}d), inheriting reliable affordance annotations while introducing geometric and appearance variation. The affordance-object co-occurrence matrix (\cref{fig:dataset_overview}e) reveals a pronounced many-to-many structure, where each object supports multiple affordances and each affordance maps to diverse categories, making intent-level disambiguation indispensable in multi-object settings.

% ============================================================
% 4. METHOD
% ============================================================
\section{Method}
\label{sec:method}

\begin{figure}[!t]
  \centering
  \includegraphics[width=\textwidth]{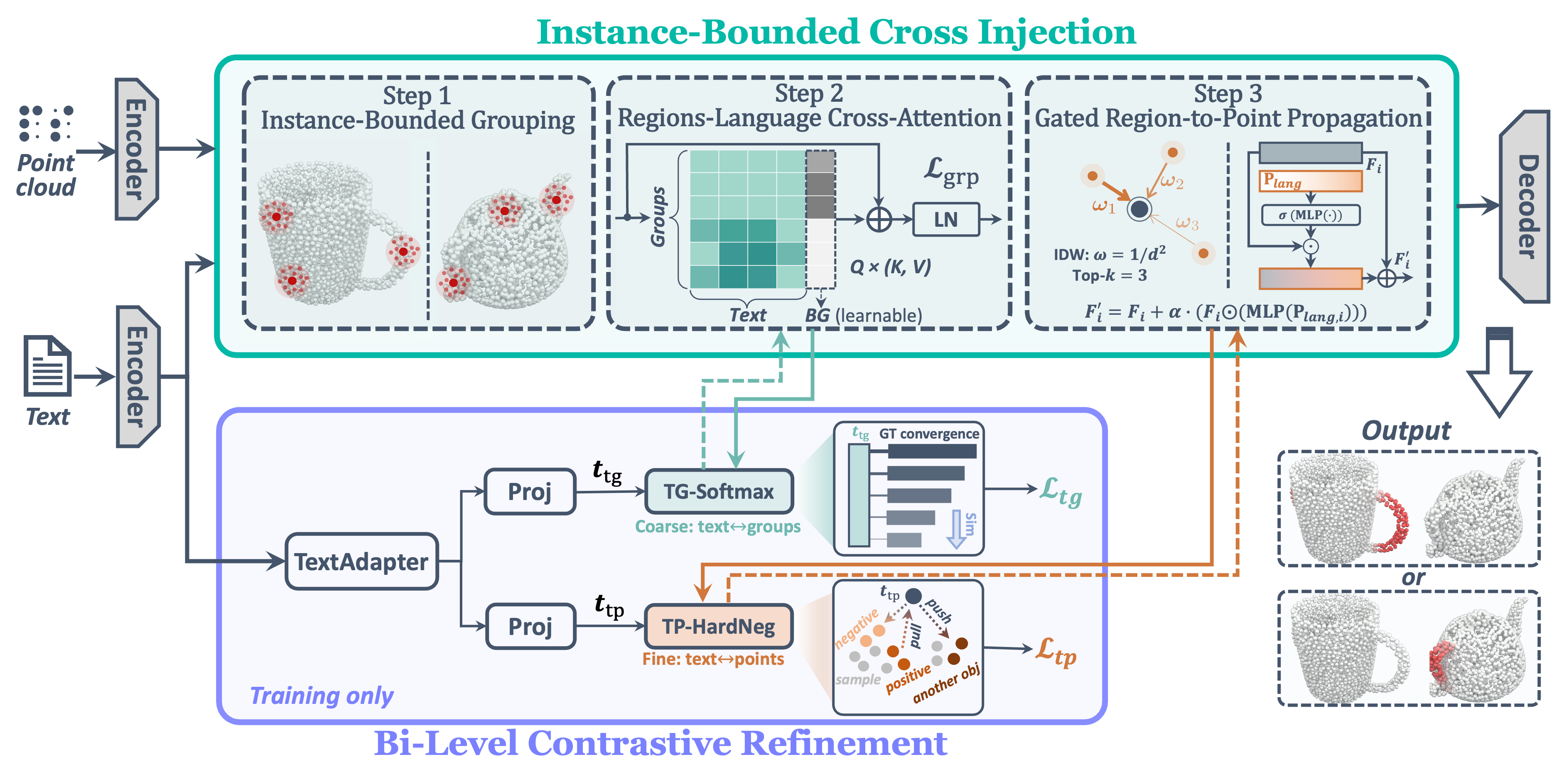}
\caption{\textbf{Overall architecture of \model{}.} \ici{} confines region--language interaction within each instance via (i) instance-bounded grouping, (ii) region-language cross-attention with a learnable background token, and (iii) gated propagation back to points. \bcr{} adds two training-only contrastive losses: TG-Softmax ranks the in-object region best matching the intent, and TP-HardNeg suppresses high-scoring negatives on confusable surfaces.}
  \label{fig:architecture}
\end{figure}

Given a multi-object composition $\mathcal{P} = \{p_i\}_{i=1}^{N}$ with $p_i \in \mathbb{R}^3$ and a query $\mathcal{Q}$, we encode them, respectively, with Uni3D~\cite{zhou2023uni3d} and RoBERTa~\cite{liu2019roberta} to obtain per-point features $\bm{F} \in \mathbb{R}^{N \times D}$ and token features $\bm{T} \in \mathbb{R}^{N_t \times D}$. \model{} is designed around the two ambiguities that arise in \ours{}: which object the query refers to among functionally competing instances, and where the affordance lies in the presence of confusable surfaces. To resolve the first, Instance-bounded Cross Injection (\ici{}, \cref{sec:ici}) groups points within each instance and confines region-language interaction to these in-object regions, so that text cues cannot cross object boundaries by construction. To resolve the second,  Bi-level Contrastive Refinement (\bcr{}, \cref{sec:bcr}) introduces two complementary discriminative objectives: a region-level loss that ranks the in-object region best matching the intent, and a point-level loss that suppresses high-scoring negatives on confusable surfaces. \bcr{} adds no parameters or computation at inference. \cref{fig:architecture} shows the overall architecture.

%The overall architecture of \model{} is shown in~\cref{fig:architecture}. Given a 3D point cloud $\mathcal{P} = \{p_i\}_{i=1}^{N} \in \R^3$ and a natural language query $\mathcal{Q}$, Uni3D~\cite{zhou2023uni3d} and RoBERTa~\cite{liu2019roberta} are adopted to obtain per-point features $\bm{F} = \{\bm{f}_i\}_{i=1}^N \in \mathbb{R}^{N \times D}$ and text features $\bm{T} \in \mathbb{R}^{N_t \times D}$, respectively.
%We then pass them to Instance-bounded Cross Injection (\ici{}, \cref{sec:ici}) for object-wise separation, avoiding the leakage of query semantics across objects. The \ici{} identifies functional regions within each object (Step 1), models the correlations among functional regions and the intent-driven instructions for enhancement (Step 2), and adaptively pulls object pixels toward the corresponding functional region enhanced by query semantics (Step 3).
%Through this coarse-to-fine paradigm, the model not only captures holistic object-query interactions for query-conditioned object selection, but also suppresses query-irrelevant point level activations to produce a semantics aware affordance map. Furthermore, we propose Bi-level Contrastive Refinement (BCR, \cref{sec:bcr}), which incorporates contrastive learning at both region and point levels to explicitly guide the model in learning multi-level object-query correlations.

\subsection{Instance-bounded cross injection}
\label{sec:ici}
Standard cross-modal fusion processes the composition as a single point set, so text cues can attach to any geometrically plausible surface, including those of competing objects that share the queried affordance. \ici{} instead enforces object boundaries as a precondition. Cross-modal attention is computed independently within each instance and \ici{} accomplishes this in three steps.

\textbf{Step 1: instance-bounded grouping.}
We run radius-graph connected components on the raw 3D coordinates to assign each point an instance label $c_i \in \{0,\ldots,K{-}1\}$, where $K$ is the number of objects. Within each instance, farthest-point sampling selects $N_g$ centres $\{\bm{\mu}_j\}$, and each centre gathers its $M$ nearest neighbours from the same instance into a region $\mathcal{G}_j$; no region therefore crosses an object boundary. The region feature $\bm{g}_j \in \mathbb{R}^D$ is obtained by mean-pooling the per-point features inside $\mathcal{G}_j$ followed by an MLP. After Step~2, a linear head $\hat{s}_j = \sigma(\bm{w}^\top \hat{\bm{g}}_j + b)$ on each cross-attended region is supervised by BCE against the soft target $\bar{y}_j = |\mathcal{G}_j|^{-1}\sum_{i\in\mathcal{G}_j} y_i$, contributing an auxiliary loss $\loss_\mathrm{grp}$.

\textbf{Step 2: region-language cross-attention.}
Within each instance, the regions $\bm{G}=[\bm{g}_1;\ldots;\bm{g}_{N_g}] \in \mathbb{R}^{N_g \times D}$ attend to the query through multi-head attention,
\begin{equation}
    \hat{\bm{G}} = \mathrm{MHA}\bigl(\bm{G},\ [\bm{T};\bm{b}_\mathrm{bg}],\ [\bm{T};\bm{b}_\mathrm{bg}]\bigr),
    \label{eq:cross_attn}
\end{equation}
where the keys and values are formed by appending a learnable background token $\bm{b}_\mathrm{bg} \in \mathbb{R}^D$ to the text features $\bm{T}$. This token serves as a sink for content that no text token can explain: under a negative query, attention can route to $\bm{b}_\mathrm{bg}$ instead of being forced onto a semantic token, which is what allows \model{} to predict an empty mask.

\textbf{Step 3: gated region-to-point propagation.}
To recover a per-point representation, each point pools from its $k_p$ nearest enhanced regions via inverse-distance interpolation in 3D space,
\begin{equation}
    \bm{f}^{\mathrm{lang}}_i = \!\!\sum_{j \in \mathcal{N}_{k_p}(i)} \!\! w_{ij}\,\hat{\bm{g}}_j,
    \quad
    w_{ij} = \frac{d_{ij}^{-1}}{\textstyle\sum_{\ell \in \mathcal{N}_{k_p}(i)} d_{i\ell}^{-1}},
    \quad
    d_{ij} = \|p_i - \bm{\mu}_j\|_2,
    \label{eq:idw}
\end{equation}
where $\mathcal{N}_{k_p}(i)$ collects the $k_p{=}3$ region centres closest to $p_i$. The aggregated signal then modulates the original point feature through a multiplicative gate,
\begin{equation}
    \bm{f}'_i = \bm{f}_i + \alpha\,\bigl(\bm{f}_i \odot \sigma(\mathrm{MLP}(\bm{f}^{\mathrm{lang}}_i))\bigr),
    \label{eq:gate}
\end{equation}
where $\sigma$ is sigmoid and $\alpha$ a learnable scalar. Because the gate enters multiplicatively, language can only re-weight channels that already exist in $\bm{f}_i$ rather than inject activations of its own, so the affordance prediction stays anchored in geometry while becoming query-aware.

\subsection{Bi-level contrastive refinement}
\label{sec:bcr}
\ici{} reduces cross-instance leakage but does not rank the in-object regions for the query, nor flag points still mistakenly activated on confusable surfaces. \bcr{} closes these gaps with two training-only contrastive losses: a region-level softmax (TG-Softmax) that ranks the in-object regions against the query, and a point-level hard-negative loss (TP-HardNeg) that suppresses confident misfired points. Both share a global text vector $\bm{t}$, masked mean-pooled from the valid tokens in $\bm{T}$, which a shared adapter then projects into anchors $\bm{t}_\mathrm{tg}, \bm{t}_\mathrm{tp} \in \mathbb{R}^D$.

TG-Softmax ranks the in-object regions by their affinity to $\bm{t}_\mathrm{tg}$, with a soft target that rewards the regions whose points are most densely affordance-positive:
\begin{equation}
    \loss_\mathrm{tg} = -\sum_{j=1}^{N_g} p_j \log \frac{\exp(\mathrm{sim}(\bm{t}_\mathrm{tg},\bm{z}_j)/\tau)}{\sum_{j'} \exp(\mathrm{sim}(\bm{t}_\mathrm{tg},\bm{z}_{j'})/\tau)},
    \quad
    p_j = \frac{\bar{y}_j^\gamma}{\sum_{j'} \bar{y}_{j'}^\gamma},
    \label{eq:tg}
\end{equation}
where $\bm{z}_j = \mathrm{MLP}(\hat{\bm{g}}_j)$ projects each region, $\mathrm{sim}$ is cosine similarity, and $\gamma{>}1$ sharpens the target. Whereas $\loss_\mathrm{grp}$ scores each region independently against its own ground truth (absolute calibration), $\loss_\mathrm{tg}$ makes regions of the same instance compete in a single softmax (relative ranking), so the two are complementary rather than redundant.

TP-HardNeg targets the points that still receive confident affordance scores despite lying on a wrong-object surface or a same-object non-affordance region -- the two main sources of point-level false positives. In each batch we sample $N_+$ affordance points $\{\bm{f}_i^+\}$ as positives and $N_-$ non-affordance points $\{\bm{f}_j^-\}$ from these two surface types as negatives, and enforce a margin $m$ between every positive and the smooth-max of the negatives:
\begin{equation}
\begin{aligned}
    \loss_\mathrm{tp} &= \frac{1}{N_+}\sum_{i=1}^{N_+} \log\!\bigl(1 + \exp\bigl(m - \mathrm{sim}(\bm{t}_\mathrm{tp}, \bm{f}_i^+) + \hat{s}^-\bigr)\bigr), \\
    \hat{s}^- &= \tau_h \log \textstyle\sum_{j=1}^{N_-} \exp\bigl(\mathrm{sim}(\bm{t}_\mathrm{tp}, \bm{f}_j^-)/\tau_h\bigr),
\end{aligned}
\label{eq:tp}
\end{equation}
where $\hat{s}^-$ smoothly approximates the hardest-negative similarity. \bcr{} is dropped at inference and adds no parameters or computation to the deployed model.

\subsection{Training objective}
\label{sec:training_obj}
The final training objective combines the standard heatmap supervision $\loss_\mathrm{hm}$~\cite{lin2017focal} with the auxiliary signal from \ici{} and the two contrastive objectives from \bcr{}:
\begin{equation}
    \loss = \loss_\mathrm{hm} + \lambda_\mathrm{grp} \loss_\mathrm{grp} + \lambda_\mathrm{tg} \loss_\mathrm{tg} + \lambda_\mathrm{tp} \loss_\mathrm{tp},
    \label{eq:total}
\end{equation}
with weights $\lambda_\mathrm{grp}, \lambda_\mathrm{tg}, \lambda_\mathrm{tp}$ balancing the auxiliary terms against $\loss_\mathrm{hm}$.

\section{Experiments}
\label{sec:experiments}

\subsection{Experimental setup}
\label{sec:setup}

\noindent\textbf{Evaluation metrics.}
Following prior works~\cite{deng2021iad,li2024laso,xu2024glance}, we report
four metrics: aIoU, AUC, SIM, and MAE, evaluated on the seen and unseen query splits to assess affordance prediction accuracy and language generalization, respectively.

\noindent\textbf{Implementation details.}
We adopt Uni3D~\cite{zhou2023uni3d} ($D{=}512$) as the point encoder and RoBERTa~\cite{liu2019roberta} as the text encoder, followed by a Transformer decoder. All models are trained for 15 epochs with batch size 96 on 2$\times$RTX PRO 6000 96\,GB GPUs using AdamW~\cite{loshchilov2019decoupled} with a cosine schedule and weight decay $10^{-3}$; module-specific learning rates and gradient clipping are listed in the appendix.

\iffalse
\noindent\textbf{Implementation Details.}
We adopt Uni3D~\cite{zhou2023uni3d} ($D{=}512$) as the point
encoder and RoBERTa~\cite{liu2019roberta} as the text encoder, followed
by a Transformer decoder.
All models are trained for 15 epochs with batch size 96 on
2$\times$RTX PRO 6000 96\,GB GPUs, using AdamW~\cite{loshchilov2019decoupled}
with a cosine learning rate schedule and weight decay $10^{-3}$.
Module-specific learning rates are set to $3{\times}10^{-4}$ (\ici{}),
$1{\times}10^{-4}$ (\bcr{}), and $1{\times}10^{-6}$ (text encoder),
with gradient clipping at 1.0 and 5.0 for backbones and auxiliary
modules, respectively.
\fi

\noindent\textbf{Baselines.}
We compare against seven representative baselines, all retrained on \ours{} with identical splits and training settings: IAGNet~\cite{luo2022grounded}, GREAT~\cite{tai2024great}, PointRefer~\cite{li2024laso}, GLANCE~\cite{xu2024glance}, 3D-SPS~\cite{luo20223d}, ReLA~\cite{liu2023gres}, and ReferTrans~\cite{li2021referring}.
We follow the baseline adaptation protocol of LASO~\cite{li2024laso} for methods originally designed for image-conditioned or 2D referring segmentation.

% ============================================================
% MAIN RESULTS
% ============================================================
\subsection{Experimental results}
\label{sec:main_results}

\noindent\textbf{Quantitative results.} As shown in \cref{tab:main}, \model{} outperforms all baselines on both splits. On \textit{Test-Seen}, it surpasses the previous state of the art GLANCE by 4.02 aIoU and achieves the best score on every metric. The advantage carries over to the harder \textit{Test-Unseen} split (+3.54 aIoU over GLANCE), confirming robust generalisation to novel objects and instructions.

\begin{table}[!ht]
\centering
\caption{\textbf{Main results on \ours{}.} All methods are trained and evaluated under the same setup. Best in bold, \underline{second-best} underlined. aIoU is the primary metric.}
\label{tab:main}
\small
\setlength{\tabcolsep}{3.0pt}
\begin{tabular}{l c cccc}
\toprule
\textbf{Method} & \textbf{Venue}
  & aIoU\,$\uparrow$ & AUC\,$\uparrow$ & SIM\,$\uparrow$ & MAE\,$\downarrow$ \\
\midrule
\multicolumn{6}{l}{\textit{Test-Seen}} \\
\midrule
3D-SPS~\cite{luo20223d} & CVPR~2022
  & 5.23 & 64.7 & 0.158 & 0.096 \\
ReferTrans~\cite{li2021referring} & NeurIPS~2021
  & 5.81 & 66.3 & 0.171 & 0.093 \\
ReLA~\cite{liu2023gres} & CVPR~2023
  & 6.47 & 68.9 & 0.193 & 0.091 \\
IAGNet~\cite{luo2022grounded} & ICCV~2023
  & 7.64 & 72.4 & 0.214 & 0.086 \\
GREAT~\cite{tai2024great} & CVPR~2025
  & 9.23 & 75.7 & 0.237 & 0.082 \\
PointRefer~\cite{li2024laso} & CVPR~2024
  & 10.52 & 79.3 & 0.260 & 0.079 \\
GLANCE~\cite{xu2024glance} & ICCV~2025
  & \underline{14.18} & \underline{87.5} & \underline{0.296} & \underline{0.077} \\
\model{} (Ours) & ---
  & \textbf{18.20} & \textbf{89.2} & \textbf{0.368} & \textbf{0.061} \\
\midrule
\multicolumn{6}{l}{\textit{Test-Unseen}} \\
\midrule
3D-SPS~\cite{luo20223d} & CVPR~2022
  & 4.12 & 61.8 & 0.136 & 0.099 \\
ReferTrans~\cite{li2021referring} & NeurIPS~2021
  & 4.58 & 63.5 & 0.148 & 0.096 \\
ReLA~\cite{liu2023gres} & CVPR~2023
  & 5.06 & 65.7 & 0.167 & 0.094 \\
IAGNet~\cite{luo2022grounded} & ICCV~2023
  & 6.07 & 68.4 & 0.186 & 0.089 \\
GREAT~\cite{tai2024great} & CVPR~2025
  & 7.31 & 72.0 & 0.209 & 0.085 \\
PointRefer~\cite{li2024laso} & CVPR~2024
  & 8.47 & 75.8 & 0.232 & 0.082 \\
GLANCE~\cite{xu2024glance} & ICCV~2025
  & \underline{11.82} & \underline{85.2} & \underline{0.268} & \underline{0.075} \\
\model{} (Ours) & ---
  & \textbf{15.36} & \textbf{87.4} & \textbf{0.339} & \textbf{0.059} \\
\bottomrule
\end{tabular}
\end{table}

\iffalse
\noindent\textbf{Quantitative Results.} As shown in \cref{tab:main}, \model{} outperforms all baselines on both \textit{Test-Seen} and \textit{Test-Unseen} settings.
On \textit{Test-Seen}, \model{} achieves an aIoU of 18.20, surpassing the previous state-of-the-art method GLANCE by 4.02. In addition, \model{} attains the highest scores across all metrics, including AUC (89.2), SIM (0.368), and the lowest MAE (0.061), indicating stronger alignment with query semantics and more precise affordance predictions.
On the more challenging \textit{Test-Unseen} split, \model{} also demonstrates significant gains, with an aIoU of 15.36, outperforming GLANCE by 3.54. Similar improvements are observed in all other metrics, highlighting the model's robust generalization to novel objects and language instructions. These results confirm the effectiveness of our method in multi-object affordance recognition and intent-driven instruction understanding, and highlight its potential applicability in real-world scenarios where unseen object instances and arbitrary instructions must be handled accurately.
\fi

% ============================================================
% QUALITATIVE ANALYSIS
% ============================================================

\begin{figure}[!ht]
\centering
\includegraphics[width=\textwidth]{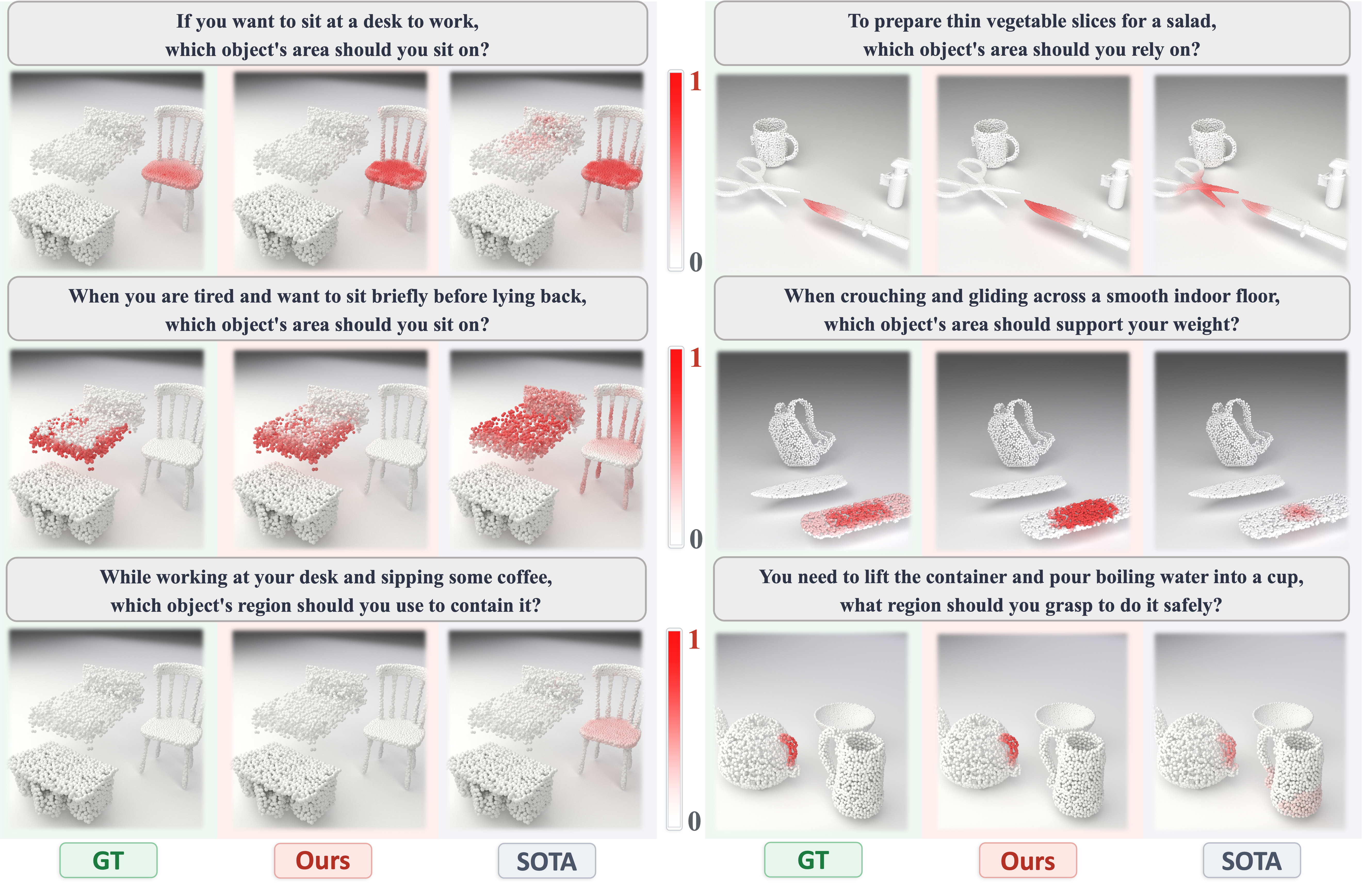}
\caption{\textbf{Qualitative comparison on \ours{}.}
Each triplet shows ground truth (GT), \model{} (Ours), and GLANCE (SOTA).
Left: the same composition queried with different intents activates different objects/regions (chair seat \vs bed surface), illustrating query-dependent disambiguation.
Right: diverse confusing pairs (knife \vs scissors, skateboard \vs surfboard, kettle \vs cup).
Red denotes higher affordance probability.}
\label{fig:qualitative}
\end{figure}

\noindent\textbf{Qualitative results.} To intuitively illustrate the capabilities of \model{}, we visualize several representative predictions on confusing compositions in \cref{fig:qualitative}.
In the left panel, different queries applied to the same composition result in distinct affordance regions (chair seat and bed surface), indicating that \model{} can generalize to diverse queries by following the corresponding semantics rather than relying on a fixed spatial prior.
In the right panel, \model{} is capable of localizing fine-grained parts, such as the knife blade for ``thin vegetable slices'' and the kettle handle for ``pour boiling water'', while suppressing the competing object with overlapping affordance.

% ============================================================
% ABLATION STUDIES
% ============================================================
\subsection{Ablation studies}
\label{sec:ablation}

\cref{tab:ablation} reports the contribution of each component. The largest drop comes from removing \ici{} ($-2.96$ aIoU), supporting our hypothesis that cross-object semantic leakage is the dominant failure mode in multi-object affordance grounding and that instance-bounded grouping is the right way to contain it. All three of its sub-components matter individually: gated propagation provides multi-level interaction, the group relevance loss steers features toward functional regions, and the background token absorbs attention from query-irrelevant points. \bcr{} adds a smaller but consistent gain on top, with TG-Softmax and TP-HardNeg accounting for drops of 0.92 and 0.58 aIoU. The two losses operate at complementary granularities: TG-Softmax aligns the correct group with the query, while TP-HardNeg sharpens point-level discrimination on the most confusable surfaces.

\begin{table}[!ht]
\centering
\caption{\textbf{Ablation study on \model{} (Seen).} Starting from the full model, we remove each module and its sub-components. All variants use identical training settings.}
\label{tab:ablation}
\small
\setlength{\tabcolsep}{3.5pt}
\renewcommand{\arraystretch}{1.05}
\begin{tabular}{l cccc}
\toprule
\textbf{Configuration}
  & aIoU\,$\uparrow$ & AUC\,$\uparrow$ & SIM\,$\uparrow$ & MAE\,$\downarrow$ \\
\midrule
\model{} (Full)
  & 18.20 & 89.2 & 0.368 & 0.061 \\
Baseline (w/o \ici{} \& \bcr{})
  & 13.80 & 87.2 & 0.290 & 0.079 \\
\midrule
w/o \ici{}
  & 15.24 & 87.8 & 0.311 & 0.073 \\
\quad w/o Background token
  & 17.58 & 88.9 & 0.360 & 0.064 \\
\quad w/o Group relevance loss
  & 17.26 & 88.6 & 0.352 & 0.066 \\
\quad w/o Gated propagation
  & 16.93 & 88.5 & 0.343 & 0.068 \\
\midrule
w/o \bcr{}
  & 16.70 & 88.4 & 0.353 & 0.066 \\
\quad w/o TG-Softmax
  & 17.28 & 88.7 & 0.357 & 0.065 \\
\quad w/o TP-HardNeg
  & 17.62 & 88.9 & 0.362 & 0.063 \\
\bottomrule
\end{tabular}
\end{table}

\iffalse
We further conduct the ablation study to show the effectiveness of key components in \model{}. The results are summarized in \cref{tab:ablation}.

\noindent\textbf{Effectiveness of \ici{}.}
As shown in \cref{tab:ablation}, removing \ici{} makes the performances drop significantly ($-2.96$ aIoU). This indicates that cross-object semantic leakage poses a fundamental challenge in multi-object affordance grounding, and our proposed \ici{} effectively mitigates this issue. Besides, each component of \ici{} contributes to performance improvements. The results demonstrate that multi-level interaction (gated propagation), guidance on functional regions feature learning (group relevance loss), and absorbing the attentions of query-irrelevant regions (background token) are essential to our tasks.

\noindent\textbf{Effectiveness of \bcr{}.}
Moreover, removing \bcr{} consistently degrades performance. Specifically, removing TG-Softmax and TP-HardNeg results in aIoU drops of 0.92 and 0.58, respectively. This fact shows that providing explicit supervision signals on functional regions that best match the queries and extremely confusing point-level affordance is effective and important. Furthermore, removing TG-Softmax and TP-HardNeg both lead to performance degradation, proving that they provide complementary supervision signals that help the model learn multi-granularity object representations more effectively.
\fi

% ============================================================
% ROBOTICS APPLICATION
% ============================================================
\subsection{Robotics application}
\label{sec:robotics}

\begin{figure}[!ht]
\centering
\includegraphics[width=\textwidth]{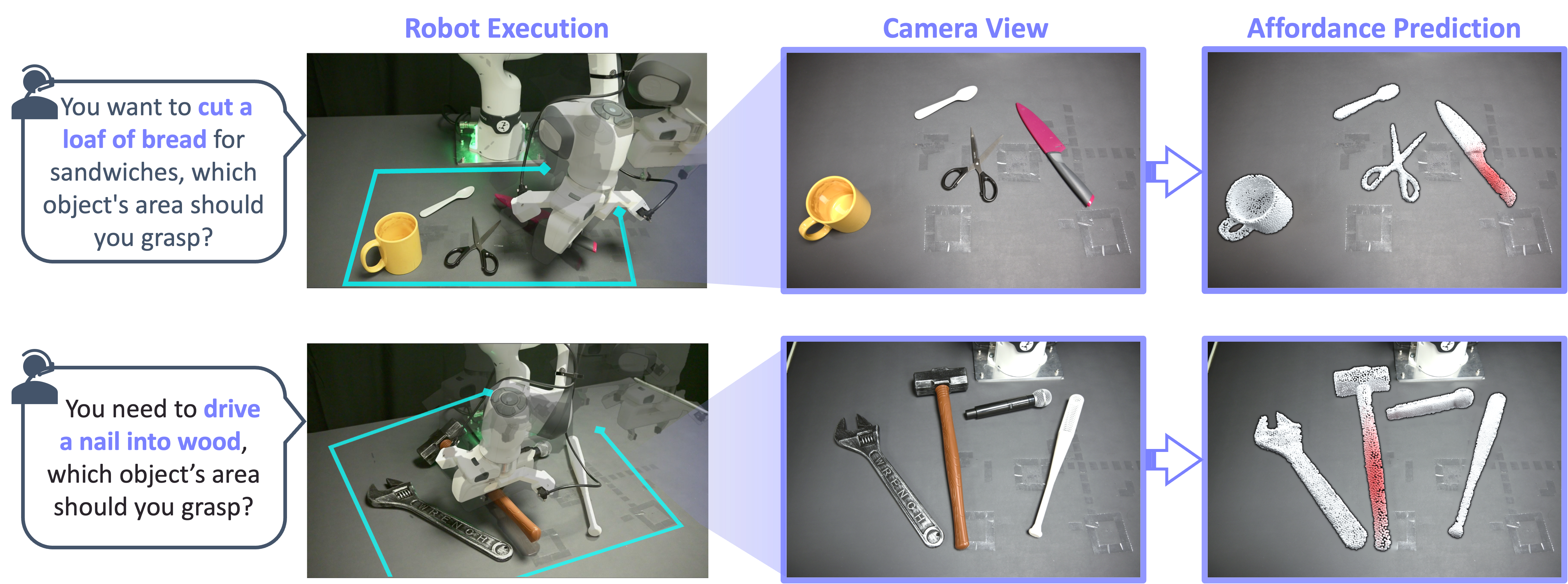}
\caption{\textbf{Real-world robotic grasping in confusing multi-object compositions.}
Each row shows the captured scene, \model{}'s affordance prediction on the reconstructed point cloud (red\,=\,high probability), and the executed grasp.
Top: a cutting query selects the knife over scissors.
Bottom: a hammering query selects the hammer over distractors.}
\label{fig:robot}
\end{figure}

Beyond controlled benchmarks, we show that \model{} generalises to real-world robotic grasping from a single RGB image.
From this input, Grounded-SAM~\cite{ren2024grounded} segments object instances using open-vocabulary prompts and Depth Anything V2~\cite{yang2024depth} estimates a dense metric depth map.
Back-projecting each masked region through a pinhole camera model yields a per-object point cloud, which we subsample to $2{,}048$ points by farthest point sampling and normalize to the unit sphere.
\model{} then takes the assembled composition together with the intent-driven instruction and predicts per-point affordances, which we map back to camera-space coordinates.
On the same point cloud, AnyGrasp~\cite{fang2023anygrasp} produces candidate grasp poses, and the final grasp is chosen by the product of grasp confidence and predicted affordance.
In the two examples of \Cref{fig:robot}, \model{} localizes the graspable handle of the intended object among confusing alternatives, despite working from partial-view pseudo point clouds.
Activations remain tightly concentrated on the correct handle and near zero elsewhere, indicating that the language query alone suffices to drive grasp selection.

\section{Conclusion and limitations}
\label{sec:conclusion}

We introduced \task{}, a setting where confusing pairs with shared affordance types make correct predictions inherently query-dependent. To study this problem, we constructed CompassAD, the first benchmark centered on implicit intent in multi-object compositions, and proposed CompassNet, which combines Instance-bounded Cross Injection (ICI) for boundary-aware cross-modal grounding with Bi-level Contrastive Refinement (BCR) for multi-granularity discrimination between target and confusable surfaces. Experiments demonstrate state-of-the-art results on both seen and unseen queries, and deployment on a robotic manipulator confirms effective real-world transfer. The main limitation is that our method assumes object instances are reasonably separable in 3D space. Under severe occlusion or tight stacking, instance-bounded grouping may span multiple objects, weakening the boundary constraints that ICI relies on. Additionally, the object and affordance categories in CompassAD are bounded by existing 3D affordance datasets. Leveraging generative 3D models to synthesize diverse compositions beyond current annotations is a promising direction for future work.

% ============================================================
% BIBLIOGRAPHY
% ============================================================
\bibliographystyle{assets/plainnat}
\bibliography{paper}

@String(ICASSP=	{ICASSP})

@inproceedings{li2025learning,
  title={Learning precise affordances from egocentric videos for robotic manipulation},
  author={Li, Gen and Tsagkas, Nikolaos and Song, Jifei and Mon-Williams, Ruaridh and Vijayakumar, Sethu and Shao, Kun and Sevilla-Lara, Laura},
  booktitle={Proceedings of the IEEE/CVF International Conference on Computer Vision},
  pages={10581--10591},
  year={2025}
}

@inproceedings{li2024one,
  title={One-shot open affordance learning with foundation models},
  author={Li, Gen and Sun, Deqing and Sevilla-Lara, Laura and Jampani, Varun},
  booktitle={Proceedings of the IEEE/CVF Conference on Computer Vision and Pattern Recognition},
  pages={3086--3096},
  year={2024}
}

@inproceedings{li2023locate,
  title={Locate: Localize and transfer object parts for weakly supervised affordance grounding},
  author={Li, Gen and Jampani, Varun and Sun, Deqing and Sevilla-Lara, Laura},
  booktitle={Proceedings of the IEEE/CVF Conference on Computer Vision and Pattern Recognition},
  pages={10922--10931},
  year={2023}
}

@article{fang2023anygrasp,
  title={Anygrasp: Robust and efficient grasp perception in spatial and temporal domains},
  author={Fang, Hao-Shu and Wang, Chenxi and Fang, Hongjie and Gou, Minghao and Liu, Jirong and Yan, Hengxu and Liu, Wenhai and Xie, Yichen and Lu, Cewu},
  journal={IEEE Transactions on Robotics},
  volume={39},
  number={5},
  pages={3929--3945},
  year={2023},
  publisher={IEEE}
}

@article{yang2024depth,
  title={Depth anything v2},
  author={Yang, Lihe and Kang, Bingyi and Huang, Zilong and Zhao, Zhen and Xu, Xiaogang and Feng, Jiashi and Zhao, Hengshuang},
  journal={Advances in Neural Information Processing Systems},
  volume={37},
  pages={21875--21911},
  year={2024}
}

@article{ren2024grounded,
  title={Grounded sam: Assembling open-world models for diverse visual tasks},
  author={Ren, Tianhe and Liu, Shilong and Zeng, Ailing and Lin, Jing and Li, Kunchang and Cao, He and Chen, Jiayu and Huang, Xinyu and Chen, Yukang and Yan, Feng and others},
  journal={arXiv preprint arXiv:2401.14159},
  year={2024}
}

@misc{gibson1979ecological,
  title={The ecological approach to visual perception},
  author={Bornstein, Marc H},
  year={1980},
  publisher={JSTOR}
}

@inproceedings{deng2021iad,
  title={3d affordancenet: A benchmark for visual object affordance understanding},
  author={Deng, Shengheng and Xu, Xun and Wu, Chaozheng and Chen, Ke and Jia, Kui},
  booktitle={proceedings of the IEEE/CVF conference on computer vision and pattern recognition},
  pages={1778--1787},
  year={2021}
}

@inproceedings{luo2022grounded,
  title={Grounding 3d object affordance from 2d interactions in images},
  author={Yang, Yuhang and Zhai, Wei and Luo, Hongchen and Cao, Yang and Luo, Jiebo and Zha, Zheng-Jun},
  booktitle={Proceedings of the IEEE/CVF International Conference on Computer Vision},
  pages={10905--10915},
  year={2023}
}

@inproceedings{li2024laso,
  title={Laso: Language-guided affordance segmentation on 3d object},
  author={Li, Yicong and Zhao, Na and Xiao, Junbin and Feng, Chun and Wang, Xiang and Chua, Tat-seng},
  booktitle={Proceedings of the IEEE/CVF Conference on Computer Vision and Pattern Recognition},
  pages={14251--14260},
  year={2024}
}

@inproceedings{xu2024glance,
  title={Intermediate Connectors and Geometric Priors for Language-Guided Affordance Segmentation on Unseen Object Categories},
  author={Li, Yicong and Chen, Yiyang and Ma, Zhenyuan and Xiao, Junbin and Wang, Xiang and Yao, Angela},
  booktitle={Proceedings of the IEEE/CVF International Conference on Computer Vision},
  pages={22836--22845},
  year={2025}
}

@article{zhou2023uni3d,
  title={Uni3d: Exploring unified 3d representation at scale},
  author={Zhou, Junsheng and Wang, Jinsheng and Ma, Baorui and Liu, Yu-Shen and Huang, Tiejun and Wang, Xinlong},
  journal={arXiv preprint arXiv:2310.06773},
  year={2023}
}

@article{liu2019roberta,
  title={Roberta: A robustly optimized bert pretraining approach},
  author={Liu, Yinhan and Ott, Myle and Goyal, Naman and Du, Jingfei and Joshi, Mandar and Chen, Danqi and Levy, Omer and Lewis, Mike and Zettlemoyer, Luke and Stoyanov, Veselin},
  journal={arXiv preprint arXiv:1907.11692},
  year={2019}
}

@inproceedings{do2018affordancenet,
  title={Affordancenet: An end-to-end deep learning approach for object affordance detection},
  author={Do, Thanh-Toan and Nguyen, Anh and Reid, Ian},
  booktitle={2018 IEEE international conference on robotics and automation (ICRA)},
  pages={5882--5889},
  year={2018},
  organization={IEEE}
}

@inproceedings{fang2018demo2vec,
  title={Demo2vec: Reasoning object affordances from online videos},
  author={Fang, Kuan and Wu, Te-Lin and Yang, Daniel and Savarese, Silvio and Lim, Joseph J},
  booktitle={Proceedings of the IEEE Conference on Computer Vision and Pattern Recognition},
  pages={2139--2147},
  year={2018}
}

@inproceedings{nagarajan2019grounded,
  title={Grounded human-object interaction hotspots from video},
  author={Nagarajan, Tushar and Feichtenhofer, Christoph and Grauman, Kristen},
  booktitle={Proceedings of the IEEE/CVF International Conference on Computer Vision},
  pages={8688--8697},
  year={2019}
}

@article{loshchilov2019decoupled,
  title={Decoupled weight decay regularization},
  author={Loshchilov, Ilya and Hutter, Frank},
  journal={arXiv preprint arXiv:1711.05101},
  year={2017}
}

@inproceedings{liu2023gres,
  title={Gres: Generalized referring expression segmentation},
  author={Liu, Chang and Ding, Henghui and Jiang, Xudong},
  booktitle={Proceedings of the IEEE/CVF conference on computer vision and pattern recognition},
  pages={23592--23601},
  year={2023}
}

@inproceedings{yang2023iagnet,
  title={ICGNet: a unified approach for instance-centric grasping},
  author={Zurbr{\"u}gg, Ren{\'e} and Liu, Yifan and Engelmann, Francis and Kumar, Suryansh and Hutter, Marco and Patil, Vaishakh and Yu, Fisher},
  booktitle={2024 IEEE International Conference on Robotics and Automation (ICRA)},
  pages={4140--4146},
  year={2024},
  organization={IEEE}
}

@article{wang2024affogato,
  title={Affogato: Learning Open-Vocabulary Affordance Grounding with Automated Data Generation at Scale},
  author={Lee, Junha and Park, Eunha and Park, Chunghyun and Kang, Dahyun and Cho, Minsu},
  journal={arXiv preprint arXiv:2506.12009},
  year={2025}
}

@inproceedings{tai2024great,
  title={Great: Geometry-intention collaborative inference for open-vocabulary 3d object affordance grounding},
  author={Shao, Yawen and Zhai, Wei and Yang, Yuhang and Luo, Hongchen and Cao, Yang and Zha, Zheng-Jun},
  booktitle={Proceedings of the Computer Vision and Pattern Recognition Conference},
  pages={17326--17336},
  year={2025}
}

@inproceedings{thermos2020deep,
  title={A deep learning approach to object affordance segmentation},
  author={Thermos, Spyridon and Daras, Petros and Potamianos, Gerasimos},
  booktitle={ICASSP 2020-2020 IEEE International Conference on Acoustics, Speech and Signal Processing (ICASSP)},
  pages={2358--2362},
  year={2020},
  organization={IEEE}
}

@inproceedings{radford2021learning,
  title={Learning transferable visual models from natural language supervision},
  author={Radford, Alec and Kim, Jong Wook and Hallacy, Chris and Ramesh, Aditya and Goh, Gabriel and Agarwal, Sandhini and Sastry, Girish and Askell, Amanda and Mishkin, Pamela and Clark, Jack and others},
  booktitle={International conference on machine learning},
  pages={8748--8763},
  year={2021},
  organization={PmLR}
}

@inproceedings{qian2024affordancellm,
  title={Affordancellm: Grounding affordance from vision language models},
  author={Qian, Shengyi and Chen, Weifeng and Bai, Min and Zhou, Xiong and Tu, Zhuowen and Li, Li Erran},
  booktitle={Proceedings of the IEEE/CVF Conference on Computer Vision and Pattern Recognition},
  pages={7587--7597},
  year={2024}
}

@article{chu20253d,
  title={3d-affordancellm: Harnessing large language models for open-vocabulary affordance detection in 3d worlds},
  author={Chu, Hengshuo and Deng, Xiang and Lv, Qi and Chen, Xiaoyang and Li, Yinchuan and Hao, Jianye and Nie, Liqiang},
  journal={arXiv preprint arXiv:2502.20041},
  year={2025}
}

@inproceedings{nguyen2023open,
  title={Open-vocabulary affordance detection in 3d point clouds},
  author={Nguyen, Toan and Vu, Minh Nhat and Vuong, An and Nguyen, Dzung and Vo, Thieu and Le, Ngan and Nguyen, Anh},
  booktitle={2023 IEEE/RSJ International Conference on Intelligent Robots and Systems (IROS)},
  pages={5692--5698},
  year={2023},
  organization={IEEE}
}

@inproceedings{chen2024worldafford,
  title={Worldafford: Affordance grounding based on natural language instructions},
  author={Chen, Changmao and Cong, Yuren and Kan, Zhen},
  booktitle={2024 IEEE 36th International Conference on Tools with Artificial Intelligence (ICTAI)},
  pages={822--828},
  year={2024},
  organization={IEEE}
}

@inproceedings{liu2025pavlm,
  title={PAVLM: Advancing point cloud based affordance understanding via vision-language model},
  author={Liu, Shang-Ching and Chen, Wenkai and Cheng, Wei-Lun and Huang, Yen-Lin and Liao, I-Bin and Li, Yung-Hui and Zhang, Jianwei and others},
  booktitle={2025 IEEE/RSJ International Conference on Intelligent Robots and Systems (IROS)},
  pages={4299--4306},
  year={2025},
  organization={IEEE}
}

@inproceedings{delitzas2024scenefun3d,
  title={Scenefun3d: Fine-grained functionality and affordance understanding in 3d scenes},
  author={Delitzas, Alexandros and Takmaz, Ayca and Tombari, Federico and Sumner, Robert and Pollefeys, Marc and Engelmann, Francis},
  booktitle={Proceedings of the IEEE/CVF Conference on Computer Vision and Pattern Recognition},
  pages={14531--14542},
  year={2024}
}

@article{xu2022partafford,
  title={Partafford: Part-level affordance discovery from 3d objects},
  author={Xu, Chao and Chen, Yixin and Wang, He and Zhu, Song-Chun and Zhu, Yixin and Huang, Siyuan},
  journal={arXiv preprint arXiv:2202.13519},
  year={2022}
}

@inproceedings{luo2022learning,
  title={Learning affordance grounding from exocentric images},
  author={Luo, Hongchen and Zhai, Wei and Zhang, Jing and Cao, Yang and Tao, Dacheng},
  booktitle={Proceedings of the IEEE/CVF conference on computer vision and pattern recognition},
  pages={2252--2261},
  year={2022}
}

@inproceedings{chen2023affordance,
  title={Affordance grounding from demonstration video to target image},
  author={Chen, Joya and Gao, Difei and Lin, Kevin Qinghong and Shou, Mike Zheng},
  booktitle={Proceedings of the IEEE/CVF Conference on Computer Vision and Pattern Recognition},
  pages={6799--6808},
  year={2023}
}

@inproceedings{bahl2023affordances,
  title={Affordances from human videos as a versatile representation for robotics},
  author={Bahl, Shikhar and Mendonca, Russell and Chen, Lili and Jain, Unnat and Pathak, Deepak},
  booktitle={Proceedings of the IEEE/CVF Conference on Computer Vision and Pattern Recognition},
  pages={13778--13790},
  year={2023}
}

@article{liu2024grounding,
  title={Grounding 3d scene affordance from egocentric interactions},
  author={Liu, Cuiyu and Zhai, Wei and Yang, Yuhang and Luo, Hongchen and Liang, Sen and Cao, Yang and Zha, Zheng-Jun},
  journal={arXiv preprint arXiv:2409.19650},
  year={2024}
}

@article{achiam2023gpt,
  title={Gpt-4 technical report},
  author={Achiam, Josh and Adler, Steven and Agarwal, Sandhini and Ahmad, Lama and Akkaya, Ilge and Aleman, Florencia Leoni and Almeida, Diogo and Altenschmidt, Janko and Altman, Sam and Anadkat, Shyamal and others},
  journal={arXiv preprint arXiv:2303.08774},
  year={2023}
}

@inproceedings{heidinger20252handedafforder,
  title={2handedafforder: Learning precise actionable bimanual affordances from human videos},
  author={Heidinger, Marvin and Jauhri, Snehal and Prasad, Vignesh and Chalvatzaki, Georgia},
  booktitle={Proceedings of the IEEE/CVF International Conference on Computer Vision},
  pages={14743--14753},
  year={2025}
}

@inproceedings{yu2025seqafford,
  title={Seqafford: Sequential 3d affordance reasoning via multimodal large language model},
  author={Yu, Chunlin and Wang, Hanqing and Shi, Ye and Luo, Haoyang and Yang, Sibei and Yu, Jingyi and Wang, Jingya},
  booktitle={Proceedings of the IEEE/CVF Conference on Computer Vision and Pattern Recognition},
  pages={1691--1701},
  year={2025}
}

@inproceedings{zhu2025grounding,
  title={Grounding 3d object affordance with language instructions, visual observations and interactions},
  author={Zhu, He and Kong, Quyu and Xu, Kechun and Xia, Xunlong and Deng, Bing and Ye, Jieping and Xiong, Rong and Wang, Yue},
  booktitle={Proceedings of the Computer Vision and Pattern Recognition Conference},
  pages={17337--17346},
  year={2025}
}

@article{jiang2025affordancesam,
  title={AffordanceSAM: Segment Anything Once More in Affordance Grounding},
  author={Jiang, Dengyang and Wang, Zanyi and Li, Hengzhuang and Dang, Sizhe and Ma, Teli and Wei, Wei and Dai, Guang and Zhang, Lei and Wang, Mengmeng},
  journal={arXiv preprint arXiv:2504.15650},
  year={2025}
}

@inproceedings{lin2017focal,
  title={Focal loss for dense object detection},
  author={Lin, Tsung-Yi and Goyal, Priya and Girshick, Ross and He, Kaiming and Doll{\'a}r, Piotr},
  booktitle={Proceedings of the IEEE international conference on computer vision},
  pages={2980--2988},
  year={2017}
}

@article{li2021referring,
  title={Referring transformer: A one-step approach to multi-task visual grounding},
  author={Li, Muchen and Sigal, Leonid},
  journal={Advances in neural information processing systems},
  volume={34},
  pages={19652--19664},
  year={2021}
}

@inproceedings{luo20223d,
  title={3d-sps: Single-stage 3d visual grounding via referred point progressive selection},
  author={Luo, Junyu and Fu, Jiahui and Kong, Xianghao and Gao, Chen and Ren, Haibing and Shen, Hao and Xia, Huaxia and Liu, Si},
  booktitle={Proceedings of the IEEE/CVF Conference on Computer Vision and Pattern Recognition},
  pages={16454--16463},
  year={2022}
}

@article{song2025learning,
  title={Learning 6-dof fine-grained grasp detection based on part affordance grounding},
  author={Song, Yaoxian and Sun, Penglei and Jin, Piaopiao and Ren, Yi and Zheng, Yu and Li, Zhixu and Chu, Xiaowen and Zhang, Yue and Li, Tiefeng and Gu, Jason},
  journal={IEEE Transactions on Automation Science and Engineering},
  year={2025},
  publisher={IEEE}
}

@inproceedings{basakvispla,
  title={ViSPLA: Visual Iterative Self-Prompting for Language-Guided 3D Affordance Learning},
  author={Basak, Hritam and Yin, Zhaozheng},
  booktitle={The Thirty-ninth Annual Conference on Neural Information Processing Systems}
}

@article{li2025seqaffordsplat,
  title={SeqAffordSplat: Scene-level Sequential Affordance Reasoning on 3D Gaussian Splatting},
  author={Li, Di and Feng, Jie and Chen, Jiahao and Dong, Weisheng and Li, Guanbin and Zheng, Yuhui and Feng, Mingtao and Shi, Guangming},
  journal={arXiv preprint arXiv:2507.23772},
  year={2025}
}

@article{wang2026videoafford,
  title={VideoAfford: Grounding 3D Affordance from Human-Object-Interaction Videos via Multimodal Large Language Model},
  author={Wang, Hanqing and Liu, Mingyu and Chen, Xiaoyu and Ma, Chengwei and Zhong, Yiming and Yin, Wenti and Liu, Yuhao and Cui, Zhiqing and Yuan, Jiahao and Dai, Lu and others},
  journal={arXiv preprint arXiv:2602.09638},
  year={2026}
}

@inproceedings{kim2024beyond,
  title={Beyond the contact: Discovering comprehensive affordance for 3d objects from pre-trained 2d diffusion models},
  author={Kim, Hyeonwoo and Han, Sookwan and Kwon, Patrick and Joo, Hanbyul},
  booktitle={European Conference on Computer Vision},
  pages={400--419},
  year={2024},
  organization={Springer}
}

@article{park2026affostruction,
  title={Affostruction: 3D Affordance Grounding with Generative Reconstruction},
  author={Park, Chunghyun and Lee, Seunghyeon and Cho, Minsu},
  journal={arXiv preprint arXiv:2601.09211},
  year={2026}
}

@article{huang2025unlocking,
  title={Unlocking 3D Affordance Segmentation with 2D Semantic Knowledge},
  author={Huang, Yu and Peng, Zelin and Wen, Changsong and Yang, Xiaokang and Shen, Wei},
  journal={arXiv preprint arXiv:2510.08316},
  year={2025}
}

@article{li2025interpretable,
  title={Interpretable Affordance Detection on 3D Point Clouds with Probabilistic Prototypes},
  author={Li, Maximilian Xiling and Rudolf, Korbinian and Blank, Nils and Lioutikov, Rudolf},
  journal={arXiv preprint arXiv:2504.18355},
  year={2025}
}

@inproceedings{wei20253daffordsplat,
  title={3daffordsplat: Efficient affordance reasoning with 3d gaussians},
  author={Wei, Zeming and Lin, Junyi and Liu, Yang and Chen, Weixing and Luo, Jingzhou and Li, Guanbin and Lin, Liang},
  booktitle={Proceedings of the 33rd ACM International Conference on Multimedia},
  pages={2821--2830},
  year={2025}
}

\end{document}